# Vision-based Situational Graphs Exploiting Fiducial Markers for the Integration of Semantic Entities

Ali Tourani[*], Hriday Bavle[†], Deniz Işınsu Avşar[‡], Jose Luis Sanchez-Lopez[†], Rafael Muñoz Salinas[§], and Holger Voos[*,¶]

June 3, 2024

**Abstract**

Situational Graphs (S-Graphs) merge geometric models of the environment generated by Simultaneous Localization and Mapping (SLAM) approaches with 3D scene graphs into a multi-layered jointly optimizable factor graph. As an advantage, S-Graphs not only offer a more comprehensive robotic situational awareness by combining geometric maps with diverse hierarchically organized semantic entities and their topological relationships within one graph, but they also lead to improved performance of localization and mapping on the SLAM level by exploiting semantic information. In this paper, we introduce a vision-based version of S-Graphs where a conventional Visual SLAM (VSLAM) system is used for low-level feature tracking and mapping. In addition, the framework exploits the potential of fiducial markers (both visible as well as our recently introduced transparent or fully invisible markers) to encode comprehensive information about environments and the objects within them. The markers aid in identifying and mapping structural-level semantic entities, including walls and doors in the environment, with reliable poses in the global reference, subsequently establishing meaningful associations with higher-level entities, including corridors and rooms. However, in addition to including semantic entities, the semantic and geometric constraints imposed by the fiducial markers are also utilized to improve the reconstructed map's quality and reduce localization errors. Experimental results on a real-world dataset collected using legged robots show that our framework excels in crafting a richer, multi-layered hierarchical map and enhances robot pose accuracy at the same time.

[*]Interdisciplinary Centre for Security, Reliability, and Trust (SnT), University of Luxembourg, Luxembourg. Institute for Advanced Studies, University of Luxembourg, Luxembourg. Corresponding author: ali.tourani@uni.lu

[†]Interdisciplinary Centre for Security, Reliability, and Trust (SnT), University of Luxembourg, Luxembourg

[‡]Institute for Advanced Studies, University of Luxembourg, Luxembourg. Department of Physics & Materials Science, University of Luxembourg, Luxembourg

[§]Department of Computer Science and Numerical Analysis, Rabanales Campus, University of Córdoba, Spain

[¶]This research was partially funded by the Luxembourg National Research Fund (FNR), DEUS Project, ref. C22/IS/17387634/DEUS; and by the Institute of Advanced Studies (IAS) of the University of Luxembourg through an "Audacity" grant (project TRANSCEND, 2021). For the purpose of open access, and in fulfillment of the obligations arising from the grant agreements, the author has applied a Creative Commons Attribution 4.0 International (CC BY 4.0) license to any Author Accepted Manuscript version arising from this submission.

# 1 Introduction

Employing vision sensors for Simultaneous Localization and Mapping (SLAM) applications in mobile robotics can bring about several merits, including the ability to achieve rich visual information using a low-cost hardware setup, making them attractive approaches compared to Light Detection And Ranging (LiDAR)-based tools [20]. These variants of SLAM systems are known as Visual SLAM (VSLAM), as they employ visual data for map reconstruction [10]. However, besides building a geometric map of the environment and being able to localize within it, advanced robotic situational awareness also requires the extraction of more abstract semantic information. To incorporate semantic data, approaches such as [4,25] enrich VSLAM with high-level information about the environment, but many of these solutions do not yet integrate valuable relational information among pertinent semantic entities. To fill this gap, 3D scene graph methodologies such as the works introduced in [2, 16, 26] exhibit promising results generating meaningful 3D scene graphs from underlying SLAM data and include dynamic, semantic, and topological relationships of the situation. While these approaches still keep the different graphs separately, the authors recently proposed as a further improvement a novel approach called Situational Graphs (*S-Graphs*) [5, 6], merging SLAM graphs and scene graphs into one multi-layered jointly optimizable factor graph with improved performance.

However, while the S-Graph approach so far only works with LiDAR sensors, this paper introduces a vision-based version by incorporating a VSLAM system. In addition, a vision-based approach also allows a simplification of the detection of particular semantic entities by exploiting visible artificial landmarks. In this regard, fiducial markers as well-known artificial landmarks for Augmented Reality (AR)/Mixed Reality (MR) or robotic tasks, such as the commonly used AprilTag [22] or ArUco [15] markers, can aid in rapid information decoding and recognition. While these conventional and often paper-based fiducial markers are visible to the human eye and hence often found to be optically distracting in many practical applications, we recently invented *iMarkers* - fiducial markers made with Cholesteric Spherical Reflectors (CSRs) [1]. The use of CSRs allows the production of mechanically very robust fiducial markers that are transparent or even shift the band of the reflected light to the UV or IR band, making them invisible for humans but still detectable for robots equipped with a specialized yet simple optical sensor. Therefore, the vision is that iMarkers will allow the future application of fiducial markers in much larger quantities in real environments without any optical distraction for humans, which finally motivates our marker-based approach in this paper.

Fiducial markers, when employed in VSLAM applications, can help to extract geometric information from the environment, and hence some VSLAM methodologies like UcoSLAM [21] and TagSLAM [23] used this potential of conventional fiducial markers to facilitate the creation of geometric maps. However, in these approaches, the performance of the localization and mapping strongly depends on the availability and detectability of the fiducial markers in the environment. Therefore, our approach leverages fiducial markers mainly to extract semantic information from the environment to create meaningful 3D scene graphs, while the SLAM level remains largely independent from the markers. In this regard, our previous preliminary work [30] partially introduced that idea, but was constrained by its exclusive reliance on monocular cameras and exhibited limited scalability in large-scale environments with various illumination conditions, providing great potential for improvement.

Therefore, this paper presents a more reliable vision-based approach combining a VSLAM system with a fiducial marker detection (both conventional markers as well as iMarkers), capable of generating a three-layered S-Graph of the environment using RGB-D cameras. It tightly couples the robot poses with structural semantic entities identified by fiducial markers, namely walls, doorways, rooms, and corridors, in a multi-layered hierarchical and optimizable S-Graph. Fig. 1 depicts an example of a generated three-layered graph for an indoor environment, interconnecting the robot poses with diverse structural and geometric priors about the scene. The principal contributions of this paper can be summarized as:

- a novel approach to create three-layered optimizable S-Graphs based on fiducial markers and supporting RGB-D visual sensors;

- a new solution for map reconstruction with a hierarchical representation procedure able to extract structural-level (*i.e.,* walls and doorways) and higher-level (*i.e.,* corridors and rooms) semantic entities,

- utilizing the potential of semantic and geometric constraints imposed by fiducial markers for improving the quality of the reconstructed map and reducing localization errors, and

- revealing the concept and potential of iMarkers for robotic situational awareness applications.

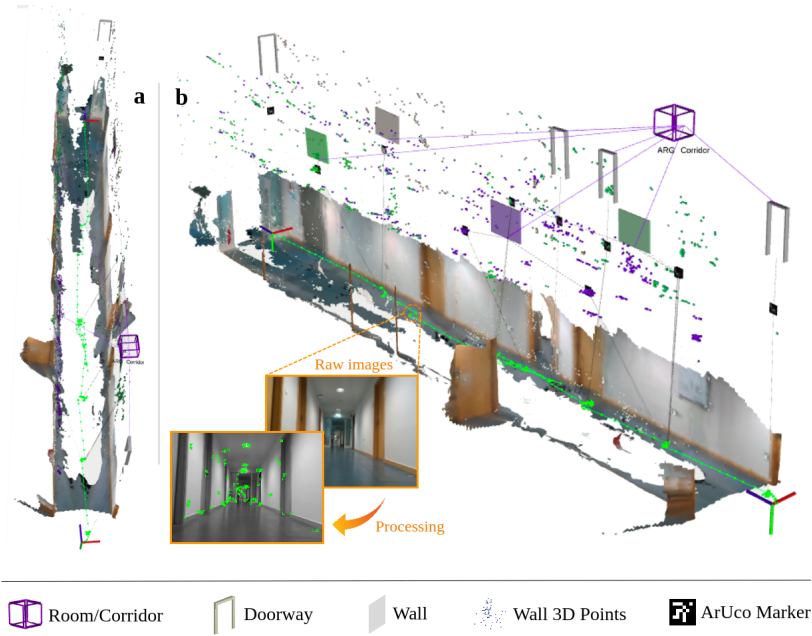

Figure 1: A reconstructed map with its hierarchical S-Graph representation generated by our framework, containing various detected structural-level entities and the connections among them: **(a)** the top view of the reconstructed map represented in 2D. **(b)** the final generated 3D view.

The rest of the paper is organized as follows: Section 2 explores previous efforts in the VSLAM domain, especially with respect to the use of structural-level data for improved reconstructed maps and the inclusion of fiducial markers. Section 3 details the proposed approach and its various modules. In Section 4, the evaluation criteria and an assessment of the effectiveness of the proposed method using real-world experimental data are provided. The paper finally concludes in Section 5.

## 2 Related Works

### 2.1 SLAM and 3D Scene Graphs

Over the years, VSLAM approaches have reached a higher level of maturity, and various innovative solutions have pushed the boundaries of this domain. The authors of this paper have surveyed the state-of-the-art in VSLAM and discussed its current trends and possible directions in [31]. According to this study, semantic VSLAM has evolved with methods such as [4,9,14] estimating a geometric map and adding semantic objects in the environment for jointly optimizing the robot pose and semantic object landmarks. Sun *et al.* [29] proposed a deep learning-based method that extracts semantic information from the scene and performs multi-object tracking based on them. However, the system does not work in real time and depends on a pre-processing step for segmentation. DS-SLAM [36] utilizes semantic information achieved by SegNet [3] for semantic mapping. The primary issue with DS-SLAM is the semantic segmentation limitations in covering various objects. PLPF-VSLAM [34] presents a VSLAM framework with an adaptive fusion of point-line-plane features, showing exciting results regardless of the richness of scene texture. Yang *et al.* [35] proposed another approach with dynamic object removal for generating static semantic maps. Although all the above methods outperform their geometric VSLAM counterparts and can classify and map different semantic elements in the environment, they can still suffer from errors due to misidentification and the semantic elements' pose estimation errors. Thus, adding structural/topological constraints among various semantic elements could further increase the robustness of the environmental understanding.

To mitigate the limitations of semantic SLAM techniques, 3D scene graph approaches such as [2,26,33] generate hierarchical representations of the environment by interconnecting different semantic entities with suitable relations. While the above techniques consider SLAM and 3D scene graphs as two different optimization problems, recent methods like [5,6,16] tightly couple SLAM graphs and 3D scene graphs for improving accuracy while generating meaningful, multi-layered hierarchical environment maps. Especially our S-Graph approach [5,6] directly merges the SLAM and the scene graph into a common multi-layered and jointly optimizable factor graph. As an advantage, these S-Graphs do not only offer a more comprehensive robotic situational awareness by combining geometric maps with diverse hierarchically organized semantic entities and their topological relationships within one graph, but they also lead to improved performance of localization and mapping on the SLAM level by exploiting semantic information. However, the generation of S-Graphs is so far limited to the exclusive use of LiDAR data.

### 2.2 Fiducial Markers and Marker-Based SLAM

While the aforementioned approaches for SLAM or semantic data extraction are based on natural visual features or objects of the environment, also fiducial markers, which are intentionally added to the environment, can serve as valuable tools for achieving enhanced scene understanding and mapping. Fiducial markers are used in robotics and Augmented Reality (AR) to encode information about objects on which they are applied, revealing to read-out devices the nature of each object and of the surrounding context. They can be classified into non-square (designed as circles [11,18,19], point sets [8,32], or arbitrary visual patterns [7,13,37]), square (or matrix-based, such as ARToolkit [17], AprilTag [22], and ArUco [15]), and hybrid variations (such as DeepTag [38] and DynaTags [27]).

However, these conventional markers are visually invasive and optically distracting if they are placed in a normal environment. Furthermore, they are often printed on pure paper which does not permit an application over a longer period of time without mechanical abrasion. Therefore, we recently introduced iMarkers which are based on an innovative optical material called Cholesteric Spherical Reflectors (CSRs) [1]. CSRs are $\sim 0.1$ mm diameter spheres of polymerized liquid crystal, exhibiting unique reflective properties. These make iMarkers invisible to humans but not to adequately designed sensors, which can detect and read them even in visually complex and dynamic environments. Designed as flexible and flat foils that can be applied to hard or soft surfaces, iMarkers exhibit omnidirectional retro-reflectivity in a narrow wavelength band with circular polarization. This enables machine detection and readout from any direction, night or day, without false positives [1]. The iMarkers can be designed to be transparent, as shown in Fig. 2, but the CSR reflection band can also be tuned outside the visible spectrum, operating in the near-infrared (IR) or near ultraviolet (UV) band and hence making the iMarkers undetectable by the human eye. Therefore, iMarkers would be non-intrusive and avoid attracting visual attention. While the iMarkers are totally novel from a material science perspective, the encoded geometric patterns we designed so far are the same as in the conventional ArUco markers, and hence, the same ArUco detection and pose estimation algorithms can also be used for the iMarkers. Further details on the optical sensors and computer vision methods we developed for the detection and readout of the iMarkers can be found in [1].

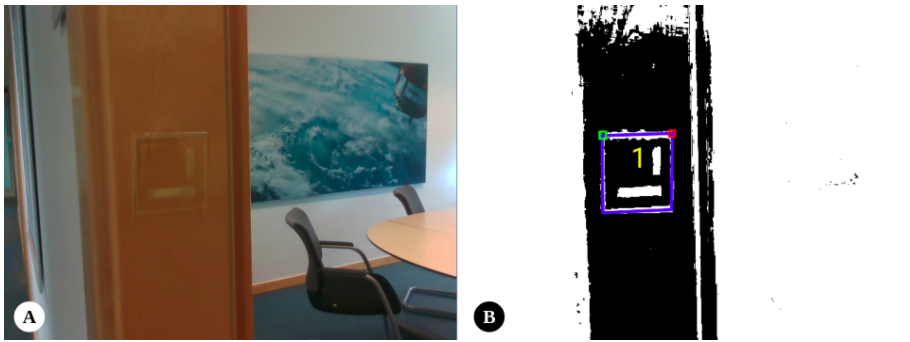

Figure 2: A transparent iMarker introduced by the authors: **(a)** an iMarker placed on a door frame captured by a normal camera, **(b)** the recognized iMarker with pose information, obtained by applying an ArUco detector on the image captured by the camera with a left-handed polarizer.

Assuming that conventional fiducial markers are placed in the environment, the literature contains several approaches that explicitly use them in VSLAM. In this regard, UcoSLAM [21] is a marker-based VSLAM solution that utilizes visual features obtained from natural landmarks and ArUco markers. It works in keypoint-only, marker-only, and mixed modes and is equipped with a marker-based loop closure detector, which requires placing distinct fiducial markers in the environment. TagSLAM [23] is another VSLAM approach that uses AprilTags to perform SLAM tasks. However, the system runs in marker-only mode and must constantly see the markers for localization and tracking stages. In another marker-based work, Romero-Ramirez *et al.* introduced a lightweight VSLAM work titled sSLAM [24], which utilizes the potential of various customized markers to facilitate tracking. It should be noted that all of the mentioned approaches are focused on creating geometric maps and marker-based loop closure detection, and they do not employ the potential of fiducial markers for decoding semantic information. Thus, the preliminary work [30] of the authors of this paper introduced the idea of using ArUco markers for the designation of semantic entities in the environment, while a modified version of UcoSLAM has then been leveraged for SLAM and the detec-tion of the marked semantic entities. However, it was limited to monocular cameras and lacked scalability in larger environments. Accordingly, the work introduced in this paper extends this preliminary approach by utilizing fiducial markers (including conventional as well as iMarkers based on the ArUco pattern) to generate a three-layered optimizable hierarchical S-Graph incorporating in particular semantic objects with appropriate relational constraints, as detailed in the following.

## 3 Proposed Method

The presented framework is built upon ORB-SLAM 3.0 [12] and aims to provide more comprehensive reconstructed maps and their multi-level topological graph representations with semantic information derived from fiducial markers. Since the first robotic applications of our proposed approach are related to buildings, *e.g.,* the autonomous monitoring of construction sites or the surveillance of indoor environments for security tasks, this paper restricts the set of included semantic entities to those that are the most important to describe the structure of building environments. Herein, walls and doorways are considered lower structural-level semantic entities that are annotated with a fiducial marker (ArUco-like conventional or iMarker), while rooms and corridors are considered higher-level semantic entities composed of the lower ones. However, the overall approach is, in principle, open to including any further semantic entities, which simply need to be annotated, in this case, with a fiducial marker and included from a mathematical point of view in the framework as described in the following.

All data are represented in the aforementioned S-Graph format [5], [6], *i.e.,* a common optimizable factor graph that merges the SLAM graph with the hierarchically ordered semantic information. Our framework computes structural-level elements' spatial positioning, leveraging fiducial markers affixed to them instead of relying on LiDAR-based odometry readings and planar surface extractions. It reconstructs a semantic map with hierarchical representations in the presence of ArUco-like markers and a database of high-level information about markers' affiliations with objects. The current version of the framework supports the RGB-D sensor, extendable to support monocular and stereo cameras. In this paper, the focus is on the generation of the S-Graph and the related algorithms, and less on the lower-level visual detection of the different included marker types, so we refer for instance to [1] regarding the sensors and visual detection of iMarkers, if present in the scene.

Fig. 3 depicts the pipeline of the proposed methodology and its constituent components. The framework benefits from a multi-thread architecture for processing data, including *tracking, local mapping, loop and map merging,* and *marker detector*. The operation commences by processing the frames captured by an RGB-D camera and conveying them to the *tracking* module where Oriented FAST and Rotated BRIEF (ORB) features and ArUco markers are extracted. The outcome of this module contains KeyFrame candidates with pose information, 3D map points, and possibly fiducial markers. If the *tracking* module decides that the current frame should be a KeyFrame, the *local mapping* module triggers to add the KeyFrame and points to the map, refining the map structure. Concurrently, the framework is being used to identify structural-level entities, including walls and doorways, and higher-level ones, containing corridors and rooms. The mentioned process takes place by leveraging pose information obtained from fiducial markers and a pre-defined *semantic perception* dictionary housing real-world data about the environment. Extracted semantic information is a resource for enhancing local bundle adjustment and KeyFrame culling the *local mapping*. The system constantly cooperates with an enhanced version of *Atlas* (the map manager module of ORB-SLAM 3.0) for establishing connections among disparate maps, representing the currently existing map (*i.e.,* active) and previously generated maps (*i.e.,* non-active). Loops and shared regions are detected within the active map and maps archived in *Atlas*, in the *loop and map merging* module. Finally, and in the case of loop detection and correction, the *global bundle adjustment* module is invoked to refine the constructed map further.

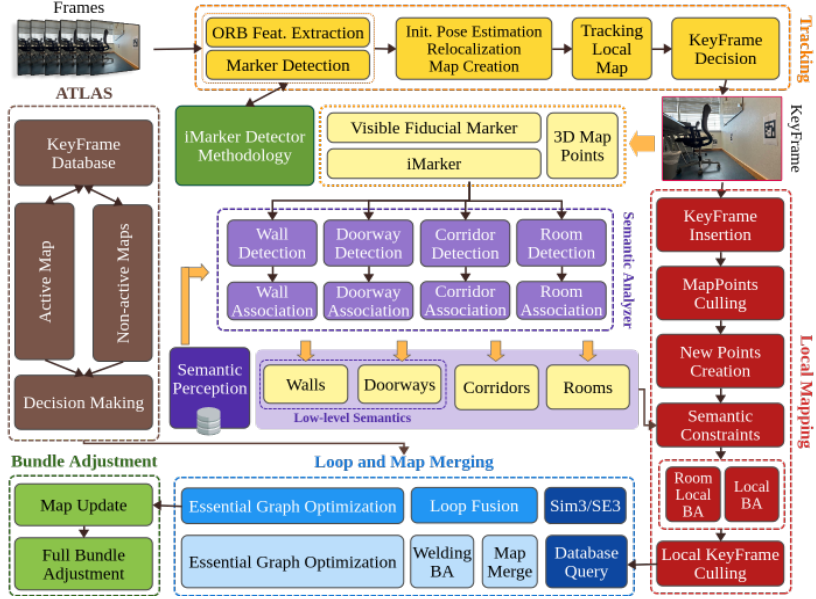

Figure 3: The primary system components and pipeline of the proposed approach.

## 3.1 Fundamentals

The proposed method introduces four central coordinate systems at time $t$: the odometry frame of reference $O$, the camera coordinate system $C_t$, the marker coordinate system $M_t$, and the global coordinate system $G_t$. The vision sensor captures a set of frames $\mathbf{F} = \{\mathbf{f}\}$, with each frame $\mathbf{f} = \{t, \mathbf{T}, \delta\}$ containing the camera pose $\mathbf{T} \in SE(3)$ acquired through the transformation of $C_t$ to $G_t$, as well as intrinsic camera parameters $\delta$. Each frame $f$ undergoes sub-sampling into an image pyramid and is processed by ORB feature extractor to obtain a set of key points for selecting KeyFrames, denoted as $\mathbf{K} = \{k\} \subset \mathbf{F}$. These KeyFrames contain feature points $\mathbf{P} = \{\mathbf{p}\}$ and fiducial markers $\mathbf{M} = \{\mathbf{m}\}$ (regardless of their type) used for detecting structural-level and semantic entities. A feature point $\mathbf{p} = \{\mathbf{x}, \mathbf{v}, \hat{\mathbf{d}}\}$ is characterized by its corresponding 3D position $\mathbf{x} \in \mathbb{R}^3$, viewing direction $\mathbf{v} \in \mathbb{R}^3$, and descriptor $\hat{\mathbf{d}}$. On the other hand, each marker $\mathbf{m} = \{id, t, s, \mathbf{p}\}$ holds unique ArUco marker identifier $m_i \in \mathbb{N}$, length $s \in \mathbb{R}$, and pose $\mathbf{p} \in SE(3)$ derived from the transformation of $M_t$ to $G_t$. Consequently, the final representation of the reconstructed map of the environment $\mathbf{E}$ is defined as follows:

$$\mathbf{E} = \{\mathbf{K}, \mathbf{P}, \mathbf{M}, \mathbf{W}, \mathbf{D}, \mathbf{R}\} \qquad (1)$$

where $\mathbf{W} = \{\mathbf{w}\}$ encompasses the detected walls within the environment, in which each wall $\mathbf{w} = \{t, \mathbf{q}, \mathbf{m_w}\}$ holds the wall equation $\mathbf{q} \in \mathbb{R}^4$ and the list of attached markers $\mathbf{m_w} \subset \mathbf{M}$. $\mathbf{D} = \{\mathbf{d}\}$ comprises the doorways found in the environment, where each doorway $\mathbf{d} = \{t, m_d, \mathbf{p}\}$ contains attached marker $\mathbf{m_d} \in \mathbf{M}$ and pose $\mathbf{p} \in SE(3)$ computed from $M_t$ to $G_t$ transformation. Finally, $\mathbf{R} = \{\mathbf{r}\}$ represents the set of rooms/corridors present in the environment, each $\mathbf{r} = \{t, \mathbf{r_c}, \mathbf{r_w}\}$ with a center point $\mathbf{r_c} \in \mathbb{R}^3$ and a list of walls $\mathbf{r_w} \subset \mathbf{W} = (w_1...w_n)|w_i \in \mathbb{N}$ that constitute the boundaries of the room or corridor.

## 3.2 Structural-level and Higher-level Semantic Entities

Reconstructing a rich semantic map of the environment incorporating the mentioned entities entails employing diverse methodologies, which will be discussed in this section. In the proposed approach, fiducial markers (which could be conventional or iMarkers) thus play a vital role, and their synergy with the system, in conjunction with the *semantic perception* module, aids in recognizing the entities of interest. Accordingly, the proposed framework takes advantage of fiducial markers, not only to add geometric constraints for map creation but also to enrich them with desired building-related semantic entities acquired with the aid of markers. It is worth emphasizing that the *semantic perception* dictionary encodes exclusively the *marker-ids* corresponding to rooms and doorways, obviating the need for any supplementary pose information of the labeled objects to be incorporated.

**Markers.** Fiducial markers are crucial reference points in our framework, enabling the system to interpret and contextualize semantic data in the environment. It has to be mentioned that the fiducial markers contain no direct information about the characteristics of the labeled items, such as pose in the global reference and width/height. Owing to their distinctive textures and the capacity to compute pose information ($\mathbf{p} \in SE(3)$), fiducial markers are primary sources of information in the proposed work for identifying and labeling targeted semantic entities, *i.e.*, walls and doorways. Each marker $\boldsymbol{m}_i$ in $G_t$ is constrained by the KeyFrame $K_i$ observing it, which can be formulated as:

$$c_{m_i}(^G\boldsymbol{K}_i, ^G\boldsymbol{m}_i) = \|^L\boldsymbol{m}_i \boxplus {}^G\boldsymbol{K}_i \boxminus {}^G\boldsymbol{m}_{i_1}\|^2_{\boldsymbol{\Lambda}_{\tilde{\boldsymbol{m}}_i}} \qquad (2)$$

where $^L\boldsymbol{m}_i$ represents the locally observed fiducial marker's pose, $\boxplus$ and $\boxminus$ refer to the composition and inverse composition, $\|\ldots\|$ is the Mahalanobis distance, and $\boldsymbol{\Lambda}_{\tilde{\boldsymbol{m}}_i}$ is information matrix associated to $\tilde{m}_i$.

**Walls.** The procedure employed to identify wall surfaces, which serve as the planar substrates on which ArUco markers and feature points features are situated, relies on pose information derived from detected markers. Wall detection occurs whenever a fiducial marker is visited within the current KeyFrame. Hence, by accessing the real-world environment data obtained from the *semantic perception* module, in cases where the marker's identifier is not found within the list of markers associated with doorways, the planar equation of the wall is calculated using the poses of the attached marker and the surrounding map points. It should be noted that this work assumes that all fiducial markers are affixed directly to the walls in an environment. Consequently, the equations characterizing these walls are obtained based on the poses of the markers attached.

Each wall $\boldsymbol{w}_i$ in $G_t$ is represented by $^G\boldsymbol{w}_i = \begin{bmatrix} ^G\boldsymbol{n}_i & ^Gd \end{bmatrix}$, where $^G\boldsymbol{n}_i = \begin{bmatrix} n_x & n_y & n_z \end{bmatrix}^T$ denotes the normal vector of the wall and $d$ represents the distance of the wall $\boldsymbol{w}_i$ from the origin in the global coordinate system. The vertex node of the wall within the graph is denoted as $^G\boldsymbol{w_i} = [^G\phi, ^G\theta, ^Gd]$, where $^G\phi$ and $^G\theta$ represent the azimuth and elevation angles of the wall in the global coordinate $G_t$, respectively. Consequently, the cost function associated with each marker $^G\boldsymbol{m}_i$ affixed to the wall $^G\boldsymbol{w}_i$ can be calculated as follows:

$$c_{w_i}(^G\boldsymbol{w}_i, ^G\boldsymbol{m}_i) = \|[^M\delta\phi_{w_{i_{m_i}}}, {}^M\delta\theta_{w_{i_{m_i}}}, {}^Md_{w_i}]^T\|^2_{\boldsymbol{\Lambda}_{\tilde{w}_i}} \qquad (3)$$

where $^M\delta\phi_{w_{i_{m_i}}}$ represents the disparity between the azimuth angle of wall $w_i$ and its attached marker $m_i$ in $M_t$, $^M\delta\theta_{w_{i_{m_i}}}$ denotes the difference in elevation angles between the two. In contrast, $^Md_{w_i}$ signifies the perpendicular distance separating the wall from the marker. This distance should ideally be zero for given marker-wall pairings.

**Corridors (Two-Wall Rooms).** Our framework leverages an adapted version of the "room segmentation" methodology originally presented in *S-Graphs+* [6], wherein markers are employed for detecting walls belonging to rooms. In this context, a corridor is defined as a room in the environment in which two parallel walls are labeled with ArUco markers. Due to the complexities associated with detecting rooms with diverse layouts, this work extends its definition to include rooms with inaccessible walls as corridors.

A corridor $^G\mathbf{r}_x = [^G\mathbf{w}_{x_{a_1}}, ^G\mathbf{w}_{x_{b_1}}]$ encompasses wall planes aligned with the $x$-axis. To calculate the center point of a corridor $^G\mathbf{r}_{x_i}$, the two equations representing the $x$-wall planes are employed in conjunction with the center point $^G\mathbf{c}_i$ of the marker $\mathbf{m}_i$ in the following manner:

$$^G\mathbf{k}_{x_i} = \frac{1}{2}|^Gd_{x_{a_1}}| \cdot {}^G\mathbf{n}_{x_{a_1}} - |^Gd_{x_{b_1}}| \cdot {}^G\mathbf{n}_{x_{b_1}} + |^Gd_{x_{b_1}}| \cdot {}^G\mathbf{n}_{x_{b_1}} \tag{4}$$

$$^G\boldsymbol{\eta}_{x_i} = {}^G\hat{\mathbf{k}}_{x_i} + {}^G\mathbf{c}_i - [\,^G\mathbf{c}_i \cdot {}^G\hat{\mathbf{k}}_{x_i}\,] \cdot {}^G\hat{\mathbf{k}}_{x_i} \tag{5}$$

where $^G\boldsymbol{\eta}_{x_i}$ represents the center point of the corridor $^G\mathbf{r}_{x_i}$ and $^G\hat{\mathbf{k}}_{x_i}$ is derived from $^G\hat{\mathbf{k}}_{x_i} = {}^G\mathbf{k}_{x_i}/\|^G\mathbf{k}_{x_i}\|$. The center point $^G\mathbf{c}_i$ of the marker is determined based on the marker's pose within the $G$ frame. Notably, the computation of the center for a two-wall room in the $y$ direction follows a similar procedure. The cost function to minimize the corridor's vertex node and its corresponding wall planes is defined as follows:

$$c_{\boldsymbol{r}_{x_i}}(^G\boldsymbol{r}_{x_i}, [^G\mathbf{w}_{x_{a_1}}, {}^G\mathbf{w}_{x_{b_1}}, {}^G\mathbf{c}_i]) = \sum_{t=1,i=1}^{T,K} \|^G\hat{\boldsymbol{\eta}}_{x_i} - f(^G\tilde{\mathbf{w}}_{x_{a_1}}, {}^G\tilde{\mathbf{w}}_{x_{b_1}}, {}^G\mathbf{c}_i)\|^2_{\boldsymbol{\Lambda}_{\tilde{r}_{i,t}}} \tag{6}$$

where $f(^G\tilde{\mathbf{w}}_{x_{a_1}}, {}^G\tilde{\mathbf{w}}_{x_{b_1}}, {}^G\mathbf{c}_i)$ is a mapping function that associates the wall planes with the corridor's center point.

**Rectangular Rooms (Four-Wall Rooms).** In the scenario where a room in the environment consists of four walls, each labeled with ArUco markers (*i.e.*, two pairs of perpendicular labeled walls), the room is represented as $^G\mathbf{r}_i = [^G\mathbf{w}_{x_{a_1}} {}^G\mathbf{w}_{x_{b_1}} {}^G\mathbf{w}_{y_{a_1}} {}^G\mathbf{w}_{y_{b_1}}]$. The center point $^G\boldsymbol{\rho}_i$ of the room $^G\mathbf{r}_i$ is computed using the following equation:

$$^G\mathbf{q}_{x_i} = \frac{1}{2}\big[|^Gd_{x_{a_1}}| \cdot {}^G\mathbf{n}_{x_{a_1}} - |^Gd_{x_{b_1}}| \cdot {}^G\mathbf{n}_{x_{b_1}}\big] + |^Gd_{x_{b_1}}| \cdot {}^G\mathbf{n}_{x_{b_1}} \tag{7}$$

$$^G\mathbf{q}_{y_i} = \frac{1}{2}\big[|^Gd_{y_{a_1}}| \cdot {}^G\mathbf{n}_{y_{a_1}} - |^Gd_{y_{b_1}}| \cdot {}^G\mathbf{n}_{y_{b_1}}\big] + |^Gd_{y_{b_1}}| \cdot {}^G\mathbf{n}_{y_{b_1}} \tag{8}$$

$$^G\boldsymbol{\rho}_i = {}^G\boldsymbol{q}_{x_i} + {}^G\boldsymbol{q}_{y_i} \tag{9}$$

where the equation is positive if $|d_{x_1}| > |d_{x_2}|$. The cost function to minimize the room's vertex node and its corresponding wall planes is defined as follows:

$$c_{\boldsymbol{\rho}}(^G\boldsymbol{\rho}, \big[^G\mathbf{w}_{x_{a_i}}, {}^G\mathbf{w}_{x_{b_i}}, {}^G\mathbf{w}_{y_{a_i}}, {}^G\mathbf{w}_{y_{b_i}}\big]) = \sum_{t=1,i=1}^{T,S} \|^G\hat{\boldsymbol{\rho}}_i - f(^G\tilde{\mathbf{w}}_{x_{a_i}}, {}^G\tilde{\mathbf{w}}_{x_{b_i}}, {}^G\tilde{\mathbf{w}}_{y_{a_i}}, {}^G\tilde{\mathbf{w}}_{y_{b_i}})\|^2_{\boldsymbol{\Lambda}_{\tilde{\rho}_{i,t}}} \tag{10}$$

where $f(^G\tilde{\mathbf{w}}_{x_{a_i}}, {}^G\tilde{\mathbf{w}}_{x_{b_i}}, {}^G\tilde{\mathbf{w}}_{y_{a_i}}, {}^G\tilde{\mathbf{w}}_{y_{b_i}})$ is a mapping function that associates wall planes with the room's center point.

**Doorways.** In this case, the pose information of ArUco markers placed on a door frame is employed to define the doorway in the map. The procedure involves visiting fiducial markers present in the current KeyFrame and verifying their association with doorways through the utilization of the *semantic perception* module. Once confirmed, the pose of the visited marker is designated for the doorway.

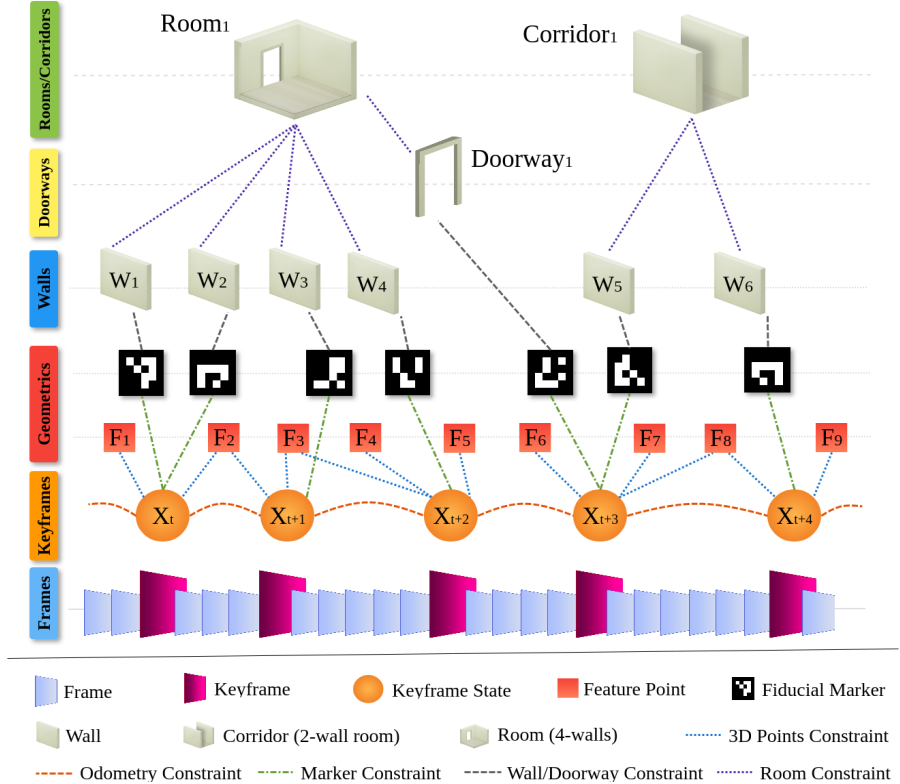

Figure 4: The hierarchical S-Graph representation of our proposed work incorporates semantic constraints, including walls, doorways, corridors, and rooms. The pre-existing geometric constraints have been complemented with semantic constraints acquired by fiducial markers.

Accordingly, the cost function for each doorway $d_i$ in $G_t$ and its corresponding room (or corridor) $^G\mathbf{r}_i$ is computed as follows [28]:

$$c_{d_i}(^G\boldsymbol{d}_i, ^G\boldsymbol{r}_i) = \|^G\hat{\boldsymbol{\delta}}_{d_i,r_j} - f(^G\boldsymbol{d}_i, ^G\boldsymbol{r}_i)\|^2_{\boldsymbol{\Lambda}_{\tilde{d}_{i,t}}} \tag{11}$$

where $^G\hat{\boldsymbol{\delta}}_{d_i,r_j}$ represents the relative distance between the door and the room and $f(^G\boldsymbol{d}_i, ^G\boldsymbol{r}_j)$ maps the relative distance among their nodes.

### 3.3 The final S-Graph

The structure of the final S-Graph generated by our framework is depicted in Fig. 4. Accordingly, the primary tracking information sources are located in the KeyFrames, which contain geometric data and pose information of objects, including 3D points and visited ArUco markers. The constraints among the mentioned objects guarantee proper computation of the odometry and loop closure detection. Structural-level entities, including walls and doorways, are linked to constraints associated with the mapped fiducial markers and establish next-level constraints with higher-level entities, including rooms and corridors.

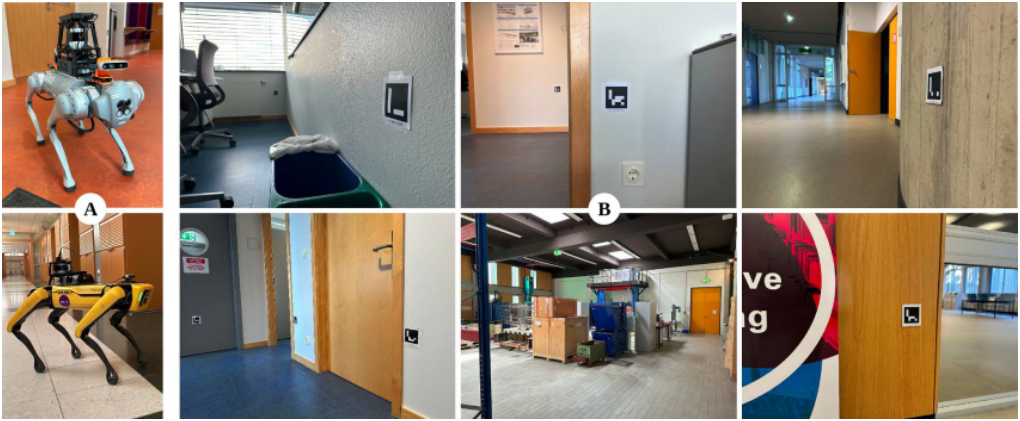

Figure 5: Dataset collected to evaluate the proposed method: **(a)** the legged robots used for data collection, **(b)** some instances of the environment prepared for data collection.

### 3.4 Inclusion of iMarkers

As conventional fiducial markers may face problems with visual clutter, they can be seamlessly replaced with iMarkers in our framework. Since the production of our iMarkers is still in an experimental stage, larger amounts of iMarkers could not yet be produced and included in the assessment. However, we applied a combination of conventional printed and a few transparent iMarkers, all with ArUco pattern, for these experiments here. The project's final future vision is to label desired semantic entities with invisible iMarkers only instead of using printed ones. The performance and usage of the mentioned markers in robotics will be evaluated in detail in Section 4.

## 4 Evaluation

To evaluate the performance and robustness of the proposed method compared to other existing frameworks, various experiments have been done using the proposed approach, UcoSLAM [21], Semantic UcoSLAM [30], and ORB-SLAM 3.0 [12] as the baseline. Due to less noisy outputs of LiDAR sensors compared to vision-based sensors, *S-Graphs+* [5, 6] as a LiDAR-based framework has been used to produce ground truth data. A computer equipped with an *11th Gen. Intel® Core™ i9 @2.60GHz* processor and 32 GigaBytes of memory was used for the evaluation mentioned.

### 4.1 Evaluation Setup

For evaluating the performance of the proposed method in real-world conditions, a 3D LiDAR sensor and an *Intel® RealSense™ Depth Camera D435i* were mounted on legged robots, including *Boston Dynamics Spot®* and *Unitree Go1* (shown in Fig. 5). The robots collected data from sensors while traversing various indoor environments with different room and corridor configurations. Each wall and door of the rooms and corridors were labeled with $8cm \times 8cm$ ArUco-like markers, and the unique identifiers of the markers were stored in a database to feed the proposed method (*i.e.*, the *semantic perception* module in Fig. 3). As previously mentioned, the environment for data collection has been prepared using a combination of printed ArUco markers and a few transparent iMarkers. Accordingly, the characteristics of the collected datasets are presented in Table 1.

Table 1: The characteristics of the collected indoors dataset.

| Sequence* | Duration | Description |
|---|---|---|
| *Seq-01* | 06m 27s | Two rooms connected via a door |
| *Seq-02* | 07m 55s | A corridor connected to a room and another corridor |
| *Seq-03* | 12m 32s | Five rooms connected to a corridor |
| *Seq-04* | 07m 34s | Two corridors connected via a landing area |
| *Seq-05* | 16m 42s | Four corridors connected to a room, forming a loop |
| *Seq-06* | 01m 44s | A single room connected to a corridor |

*data were stored as packages of *rosbag* files.

## 4.2 Experimental Results

Table 2: Evaluation results on the collected dataset using Root Mean Square Error (RMSE) error in *meters* and Standard Deviation (STD). The best results are boldfaced and the second best are underlined. Our method outperforms the state-of-the-art in most of the sequences.

|  | RMSE | | | | | | STD | | | | |
|---|---|---|---|---|---|---|---|---|---|---|---|
|  | *Seq-01* | *Seq-02* | *Seq-03* | *Seq-04* | *Seq-05* | *Seq-06* | *Seq-01* | *Seq-02* | *Seq-03* | *Seq-04* | *Seq-05* |
| Proposed | **0.5127** | 0.6662 | **2.3555** | **0.4479** | 2.1794 | **0.2189** | **0.2454** | **0.3332** | **0.7441** | **0.2422** | **0.7107** |
| *UcoSLAM [21]* | 5.7996 | 3.0521 | 3.3034 | 2.1573 | 15.0184 | 1.5601 | 3.1814 | 1.3999 | 1.2332 | 1.2284 | 6.1595 |
| *ORB-SLAM3 [12]* | 0.5351 | **0.6484** | 2.5011 | 0.4895 | **2.1404** | 0.2479 | 0.2572 | 0.3334 | 0.8602 | 0.2653 | 0.7366 |
| *Previous [30]* | 4.9437 | 2.8363 | 2.5154 | 1.9154 | 4.6672 | 1.5552 | 2.7065 | 1.3191 | 0.8582 | 1.1547 | 2.3891 |

**Accuracy.** To validate the accuracy of the proposed method in comparison to its baseline and ground truth, Absolute Trajectory Error (ATE) measurements have been employed in this paper. Regarding the evaluation results presented in Table 2, it becomes evident that our approach outperforms its baseline counterpart and other frameworks across various scenarios. This enhancement can be attributed to the ability of the proposed method to introduce new constraints to the map by associating structural-level and semantic entities. It is noteworthy that the improvements are particularly pronounced in comparison to the two marker-based methodologies. While on *Seq-02* and *Seq-05* ORB-SLAM 3.0 exhibits slightly better performance than our method, the difference is negligible (*i.e.,* < 3*cm*). This discrepancy could be due to the noisy detection of the fiducial markers as the primary source of semantic information in real-world scenarios with changing light conditions. However, it is important to emphasize that the proposed approach, in addition to improving ATE in most cases, can generate a three-layered situational graph of the environment. Accordingly, Fig. 6 depicts some qualitative results alongside the accuracy of the proposed approach against the LiDAR-based benchmark.

**The performance of the iMarkers.** To verify the applicability and potential of our described iMarkers for the first time in robotics, the door frames in *Seq-06* were labeled with a prototype of transparent iMarkers. The reconstructed map using the iMarkers is depicted in Fig. 6. Given that the sole distinction between iMarkers and conventional printed ArUco markers lies in the detection step, the resulting semantic map remained unaltered. The authors believe this will provide proper room for future investigation on utilizing such iMarkers in their future works.

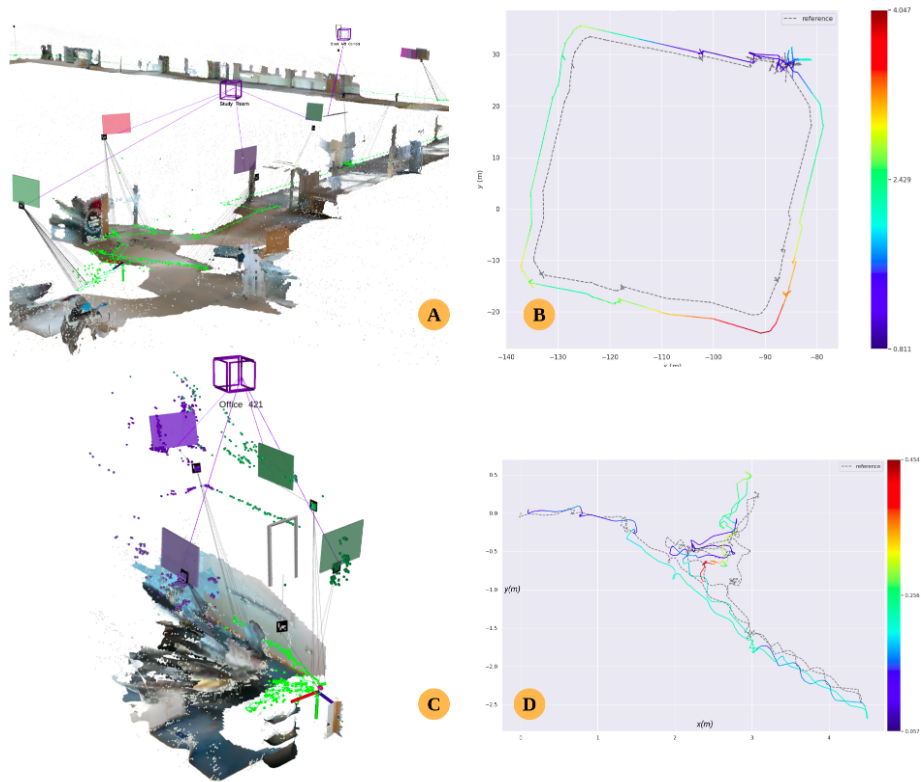

Figure 6: The qualitative results and Absolute Trajectory Error (ATE) of the proposed approach w.r.t. translation in *meters* on *Seq-01* (A and B) and *Seq-06* (C and D). The dotted lines in the charts are LiDAR ground truth values.

## 5 Conclusions

This paper introduced a VSLAM framework that effectively harnesses the outputs provided by RGB-D cameras to achieve highly accurate map reconstruction in the generation of S-Graphs. The proposed approach employs pose and topological information derived from strategically positioned ArUco markers, both conventional printed markers, and our invented iMarkers, within indoor environments to detect semantic objects, including walls, doorways, corridors, and rooms. It utilizes the added semantic entities and the topological constraints among them for an elevated reconstructed map quality. Considering the evaluations performed on a real-world dataset collected by legged robots and benchmarked against a LiDAR-based framework as the ground truth, the proposed method showed an accuracy and performance improvement compared to state-of-the-art works.

As the proposed framework is part of a broader research project, in future works, the authors intend to determine transparent objects (*e.g.,* windows, mirrors, and glass doors) using the iMarkers due to their challenging recognition using computer vision algorithms and potential difficulties for robots performing SLAM tasks. Moreover, supporting more visual (*i.e.,* mono and stereo cameras) and inertial (*i.e.,* Inertial Measurement Unit (IMU)) sensors along with the efficient implementation of modules is another target of future works. The authors also plan another extension of their work to reserve fiducial markers for higher-level tasks (*i.e.,* incorporating specific semantic information of rooms and corridors) instead of utilizing them for structural-level semantic object detection, which can be replaced by scene semantic segmentation.